\documentclass[twocolumn]{IEEEtran}
\IEEEoverridecommandlockouts
\usepackage{cite}
\usepackage{amsmath,amssymb,amsfonts}
\usepackage{algorithm}
\usepackage{algorithmic}
\usepackage{booktabs}
\usepackage{graphicx}
\usepackage{textcomp}
\usepackage{xcolor}
\usepackage{subfigure}
\newtheorem{definition}{Definition}
\newtheorem{lemma}{Lemma}
\newtheorem{proposition}{Proposition}
\newtheorem{remark}{Remark}
\newtheorem{assumption}{Assumption}
\begin{document}

\title{Robust Federated Learning with Noisy Communication}

\author{Fan Ang,
        Li Chen,
        Nan Zhao,~\IEEEmembership{Senior Member,~IEEE,}
        Yunfei Chen,~\IEEEmembership{Senior Member,~IEEE,}
        Weidong Wang,

\and F. Richard Yu,~\IEEEmembership{Fellow,~IEEE}
\thanks{

F. Ang, L. Chen and W. Wang are with Department of Electronic Engineering and Information Science, University of Science and Technology of China.(e-mail: angfan@mail.ustc.edu.cn, \{chenli87, wdwang\}@ustc.edu.cn).

N. Zhao is with the School of Info. and Commun. Eng., Dalian University of Technology, Dalian 116024 China, and also with National Mobile Communications Research Laboratory, Southeast University, Nanjing 210096, China (email: zhaonan@dlut.edu.cn).

Y. Chen is with the School of Engineering, University of Warwick, Coventry
CV4 7AL, U.K. (e-mail: Yunfei.Chen@warwick.ac.uk).

F.R. Yu is with the Department of Systems and Computer Engineering, Carleton University, Ottawa, ON, K1S 5B6, Canada (email: richard.yu@carleton.ca).
\vspace{1.5em}
}
}


\maketitle

\begin{abstract}
Federated learning is a communication-efficient training process that alternates between local training at the edge devices and averaging the updated local model at the central server.
Nevertheless, it is impractical to achieve a perfect acquisition of the local models in wireless communication due to noise, which also brings serious effects on federated learning.
To tackle this challenge, we propose a robust design for federated learning to alleviate the effects of noise in this paper.
Considering noise in the two aforementioned steps, we first formulate the training problem as a parallel optimization for each node under the expectation-based model and the worst-case model.
Due to the non-convexity of the problem, a regularization for the loss function approximation method is proposed to make it  tractable.
Regarding the worst-case model, we develop a feasible training scheme which utilizes the sampling-based successive convex approximation algorithm to tackle the unavailable maxima or minima noise condition and the non-convex issue of the objective function.
Furthermore, the convergence rates of both new designs are analyzed from a theoretical point of view.
Finally, the improvement of prediction accuracy and the reduction of loss function are demonstrated via simulations for the proposed designs.
\end{abstract}


\section{Introduction}
\label{sec:introduction}
\IEEEPARstart{F}{uture} wireless computing applications demand higher bandwidth, lower latency and more reliable connections with numerous devices \cite{li20185g}.
With the burgeoning development of artificial intelligence technologies, the edge devices need to generate a sheer volume of raw data to be transmitted to the center, which results in excessive latency and privacy concerns \cite{zhang2019deep,lee2019deep}.
To solve this problem, federated learning has been proposed to encounter a paradigm shift from computing at the center to computing at the edge devices \cite{mcmahan2016communication}.

Federated learning can be traced back as federated optimization to decouple the data acquisition and computation at the central server \cite{konevcny2015federated}.
Federated optimization has recently been extended to deep learning platforms, which was known as federated learning \cite{mcmahan2016communication, konevcny2016federated}.
Federated learning was designed as an iterative process between distributed learning at the edge devices and averaging the updated local models at the central server.
In contrast to the conventional centralized training, federated learning is more efficient in communication by uploading no raw data but only local models.
To further improve the availability of enormous data from edge devices, federated learning was adopted in several scenes of future wireless networks \cite{wang2019adaptive, samarakoon2018federated, wang2019edge, bennis2018ultrareliable}.
Using federated learning and distributed MEC systems, the authors studied the trade-off between local computing and global aggregation under the given resource-constrained model in \cite{wang2019adaptive}.
Moreover, the attractive property of lower latency drew attention to exploiting federated learning in latency-sensitive networks, such as vehicular networks \cite{samarakoon2018federated, bennis2018ultrareliable}.

Due to the high-dimensional local model, as well as the long-term training process, the updating step of federated learning still consumes a lot of communication resources.
The key issues are to reduce the overhead in the updating steps and to accelerate the training process.
A series of research concentrating on reducing the overhead in the updating step was to transmit the compressed gradient vector via exploiting the quantization scheme \cite{aji2017sparse, lin2017deep}.
Another research focused on scheduling the edge devices to save the transmission bandwidth \cite{yang2019scheduling, chen2018lag, chen2016revisiting, kamp2018efficient, nishio2019client}.
Specifically, some novel updating rules were worked out, which only allowed the edge devices with significant training improvement\cite{chen2018lag}, or the fast responding devices \cite{chen2016revisiting}, to transmit their gradient vectors in each uploading round.
Arranging the adaptive maximum number of transmission-permitted edge devices was also an intelligent way when time was limited \cite{nishio2019client}.
Furthermore, the authors developed a momentum method and cp-stochastic gradient descent algorithm to accelerate the training process for each edge device in local training in \cite{lin2017deep, agarwal2018cpsgd}.
Utilizing the different computation capability of each node, an asynchronous federated learning scheme was proposed to reduce the training delay in \cite{sprague2018asynchronous}.

The aforementioned pioneering works are all based on the assumption that the received signals at both the central server and the edge nodes are perfectly detected.
In practice, this is difficult in wireless communications due to imperfect channel estimation, feedback quantization, or delay in signal acquisition on fading channels.
In other words, the noise is indispensable during the training process.
Furthermore, neural networks were proved to be not very robust to noise, which leads to the delay in the training process \cite{tang2010deep}.

In conventional centralized learning, a branch of research has been dedicated to eliminate the effects of noise, among which several works used  the denoising autoencoder to filter noise, such as contractive auto-encoders and denoising auto-encoders \cite{rifai2011contractive, vincent2008extracting}, while others considered representing the effect of noise  as imposing a penalty during the training process, known as the regularization scheme \cite{bishop1995regularization, bishop1995training, graves2011practical, hochreiter1995simplifying, srivastava2014dropout}.
In particular, the addition of noise with infinitesimal variance as the input of training dataset was proved to be equivalent to the punishment on the norm of the weights for some training models \cite{bishop1995regularization, bishop1995training}, whereas the added noise in the model was derived as appending a regularizer in the loss function which pushes the model to find the minima in the flat regions \cite{graves2011practical, hochreiter1995simplifying}.
Besides, the key idea of the Dropout method is to randomly drop units from the neural network during training to simulate the regularization \cite{srivastava2014dropout}.
However, to the best of our knowledge, no noise reduction has been studied for  federated learning and it is still an open problem.

Motivated by these observations, we propose a robust federated learning method to alleviate the effects of noise in the training process.
Robust designs are first introduced using the expectation-based model and the worst-case model. More specifically, the former model is based on the statistical properties of the noise uncertainty and the latter model represents the fixed uncertainty sets of noise.
Furthermore, the corresponding convergence analysis is provided to illustrate the performance of the proposed designs.
The main contributions of this work are summarized as follows.

\begin{itemize}

\item[.]{\bf{Robust design under the expectation-based model.}} With the consideration of noise at the central server  and the edge nodes, we formulate the training problem using the expectation-based model as a parallel optimization problems for each edge node. To handle the statistical property of noise, as well as the non-convexity of the objective function, we propose a regularization for loss function approximation (RLA) algorithm to approach the objective function and develop the corresponding training process. The proposed solution is superior to the conventional scheme that ignores noise in terms of both prediction accuracy and performance of loss function.

\item[.]{\bf{Robust design under the worst-case model.}} The training problem under the worst-case model meets the challenges that are the unavailable maxima or minima noise condition and the non-convex issue of the objective function. We solve the former problem via the sampling method and tackle the latter one  by utilizing the successive convex approximation (SCA) algorithm to generate a feasible descent direction for the training process. The simulation results show that the proposed design outperforms the conventional one for prediction accuracy and values of loss function.

\item[.]{\bf{Convergence analysis for the proposed designs.}} The convergent property of all proposed designs are derived. Specifically, it is found that the proposed training process under the expectation-based model converges at the equivalent rate to the centralized training scheme that ignores noise, and the convergent property of proposed robust design under the worst-case model outperforms the conventional centralized one.
\end{itemize}

The remainder of the paper is organized as follows. Section II introduces the system model of the federated learning considering noise.
Section III presents the formulated problem under the expectation-based model and the worst-case model.
The robust design under the expectation-based model and its convergence analysis are developed in Section IV.
Section V shows the robust design under the worst-case model and the corresponding convergence analysis.
Simulation results are provided in Section VI.

Throughout the paper, we use boldface lowercase to refer to vectors, and lowercase to refer to scalar.
Let ${(\cdot)^T}$ denote the transpose of a vector.
Let ${|\cdot|}$ denote size of the set, ${\bf{0}}$ denotes zero matrix, and ${\bf{I}}$ denotes unit matrix. ${\mathbb{E}}\left\{\cdot\right\}$ is the expectation function.

%

\section{System Model}
\label{sec:System Model}

\begin{figure}
  \centering
  \includegraphics[width=0.5\textwidth]{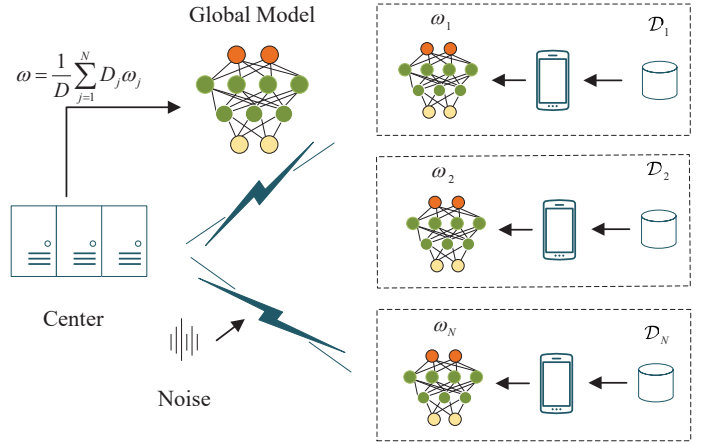}
  \caption{Federated learning with wireless communication.}
  \label{federated learning}
\end{figure}

We consider a distributed learning system consisting of a single central server and $N$ edge nodes, as shown in Fig. \ref{federated learning}.
A shared learning process with the global model ${\bf{w}}$ is trained collaboratively by the edge nodes.
Each node collects a fraction of labelled training datasets $\mathcal{D}_1, \mathcal{D}_2,..., \mathcal{D}_N$.

The loss function is to facilitate the learning and we define it as $f_j ({\bf{{\bf{w}}}}, {\bf{x}}_j, y_j)$ for each data sample $j$, which consists of the input vector ${\bf{x}}_j$ and the output scalar $y_j$. For convenience, we rewrite $f_j ({\bf{w}}, {\bf{x}}_j, y_j)$ as $f_j ({\bf{{\bf{w}}}})$.
Then the global loss function on all distributed datasets can be defined as

\begin{equation}
\label{lfunction_global}
F({\bf{w}}) = \frac {\sum_{j\in {\mathcal{D}}}f_ j({\bf{w}})}{|\cup_i \mathcal{D}_i|},\\[2mm]
\end{equation}
where $|\cdot|$ denotes the size of the datasets and each dataset $\mathcal{D}_i$ satisfies $\mathcal{D}_i \cap \mathcal{D}_j = \emptyset$ when $ i \neq j$, $i,j = 1, 2,..., N$.
The training target is to minimize the global loss function $F({\bf{w}})$ according to the distributed learning, i.e., to find

\begin{equation}
\label{w_optimize_global}
{\bf{w}}^* = \arg \min F({\bf{w}}).\\[2mm]
\end{equation}

One way to search for the optimal ${\bf{w}}^*$ is to update the datasets of the distributed nodes, which only contains the input vector ${\bf{x}}_j$ and the output scalar $y_j$, called centralized learning.  The center completes the training process using the whole datasets, and broadcasts the optimal model from (\ref{lfunction_global}) and (\ref{w_optimize_global}) to all nodes.
However, the datasets are generally large in machine learning.
Therefore, centralized learning requires numerous communication resources to collect the whole datasets.
In other words, the training process will be limited by the communication rates.

Another way to solve (\ref{w_optimize_global}) is a distributed manner as demonstrated in Fig. {\ref{federated learning}}, which focuses on the model-averaging for the global model $\bf{{\bf{w}}}$, called federated learning.
The global loss function $F({\bf{w}})$ cannot be directly computed without sharing datasets among all edge nodes in federated learning.
The federated learning algorithm alternates between two stages.
In the first stage, the local models at each node are sent to the center for model-averaging via  wireless links, and the center updates the global model ${\bf{w}}$.
In the second stage, the center broadcasts the current model to all edge nodes at each iteration. Based on the received global model ${\bf{w}}$, each node updates its own model to minimize the local loss function using its own dataset. The updating rules follow:

\begin{subequations}
\label{ww}
\begin{align}
\label{w_aggregation}
&\textup{Center} : {\bf{w}} = \frac{\sum_{j=1}^N D_j {\bf{w}}_ j}{D},\\[2mm]\label{w_optimize_local}
&\textup{Local} : {\bf{w}}_j = \arg \min F_j({\bf{w}}), j = 1, 2, ..., N,
\end{align}
\end{subequations}\\[0.5mm]
where ${\bf{w}}_ j$ denotes the local model of node $j$, $D$ denotes the size of the whole datasets ${\cup_j \mathcal{D}_j}$, $D_j$ denotes the size of the dataset $\mathcal{D}_j$, $ j = 1, 2, ..., N$, $F_j({\bf{w}})$ is the local loss function of node $j$ with dataset $\mathcal{D}_j$, and can be written as

\begin{equation}
\label{lfunction_i}
F_ j({\bf{w}}) = \frac{1}{|\mathcal{D}_j|}\sum_{i\in D_j} f_i ({\bf{w}}) = \frac{1}{D_j}\sum_{i\in {\mathcal{D}_j}} f_i ({\bf{w}}).\\[2mm]
\end{equation}
The training process requires the iterations between (\ref{w_optimize_local}) and (\ref{w_aggregation}) until convergence, and each node can obtain the optimal model ${\bf{w}}^*$.

%
Since the  center and each node are connected using wireless links, it inevitably introduces noise.
Therefore, the received signal has the aggregation noise at the center via local updating and the broadcasted global model with noise in each iteration for the node $j$ can be modeled as

\begin{equation}
\label{noise}
\begin{aligned}
\textup{Aggregation}: &\,\,\, \tilde{{\bf{w}}} = {\bf{w}} + \Delta \tilde{{\bf{w}}},\\[2mm]
\textup{Broadcast}: &\,\,\, \tilde{{\bf{w}}}_j = {\tilde{\bf{w}}} + \Delta \tilde{{\bf{w}}}_j, j = 1, 2, ..., N,\\[2mm]
\end{aligned}
\end{equation}
where $\Delta \tilde{{\bf{w}}}$ refers to the aggregation noise at the center, and $\Delta \tilde{{\bf{w}}}_j$ refers to the broadcast noise for node $j$.

The imperfect estimation is a major problem in wireless communication.
In federated learning, it leads to the changing of optimization in the local update process.
The noise in estimation error of the model will make the output data point blurred and make the training difficult to fit the input data point precisely for neural networks.
Furthermore, the neural networks were proved to be not robust against noise.
In other words, the performance of the learning scheme may be significantly reduced by noise.
To solve this problem, robust design is proposed to ensure a certain level of the performance under the uncertainty model.

\section{Problem Formulation}
\label{sec:Problem Formulation}
In this section, we formulate the robust problem using two robust models.
According to the different characteristics of the two robust models, the corresponding problem is totally different.
We write the corresponding problems in the following.

The aggregation noise and broadcasted noise in (\ref{noise}) can be modelled as the stochastic and the deterministic.
The former is the expectation-based model and the latter is the worst-case model.
According to that, each node updates its own model with a different initial point, $\tilde{{\bf{w}}}_j$, the corresponding local loss function $F_j({\bf{w}})$ is rewritten as $F_j(\tilde{{\bf{w}}}_j)$, $j = 1, 2, ..., N$, and the global loss function $F({\bf{w}})$ is rewritten as $F(\tilde{{\bf{w}}})$.
The iteration process still follows (\ref{w_aggregation}) and (\ref{w_optimize_local}).

\subsection{Training Under Expectation-based Model}
Expectation-based model is a stochastic method to represent the random condition, which can only be used when statistical properties of noise are available {\cite{ang2019confer}}.
The stochastic model assumes that the estimation value is a random quantity and its instantaneous value is unknown, but its statistics property, such as the mean and the covariance, is available.
In this case, the robust design usually aims at optimizing either the long-term average performance or the outage performance. The corresponding robust model is called the expectation-based model and defined as follows.

\begin{figure}
\centering
\subfigure[Noise under expectation-based model in two-dimensional space.]{
\label{fig:subfig:a}
\includegraphics[width=0.22\textwidth]{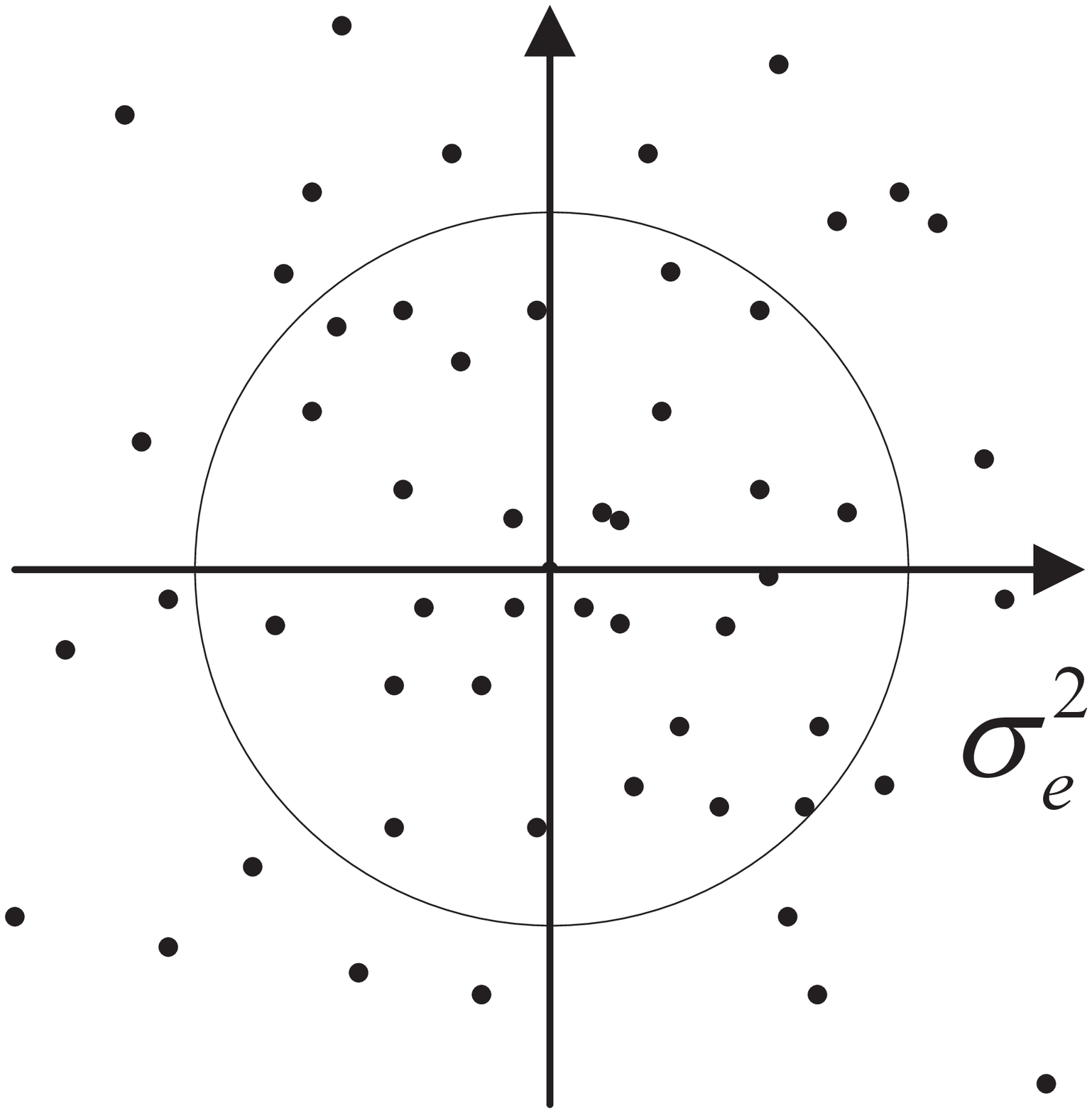}}
\hspace{0.1in} \subfigure[Noise under worst-case model in two-dimensional space.]{
\label{fig:subfig:b}
\includegraphics[width=0.22\textwidth]{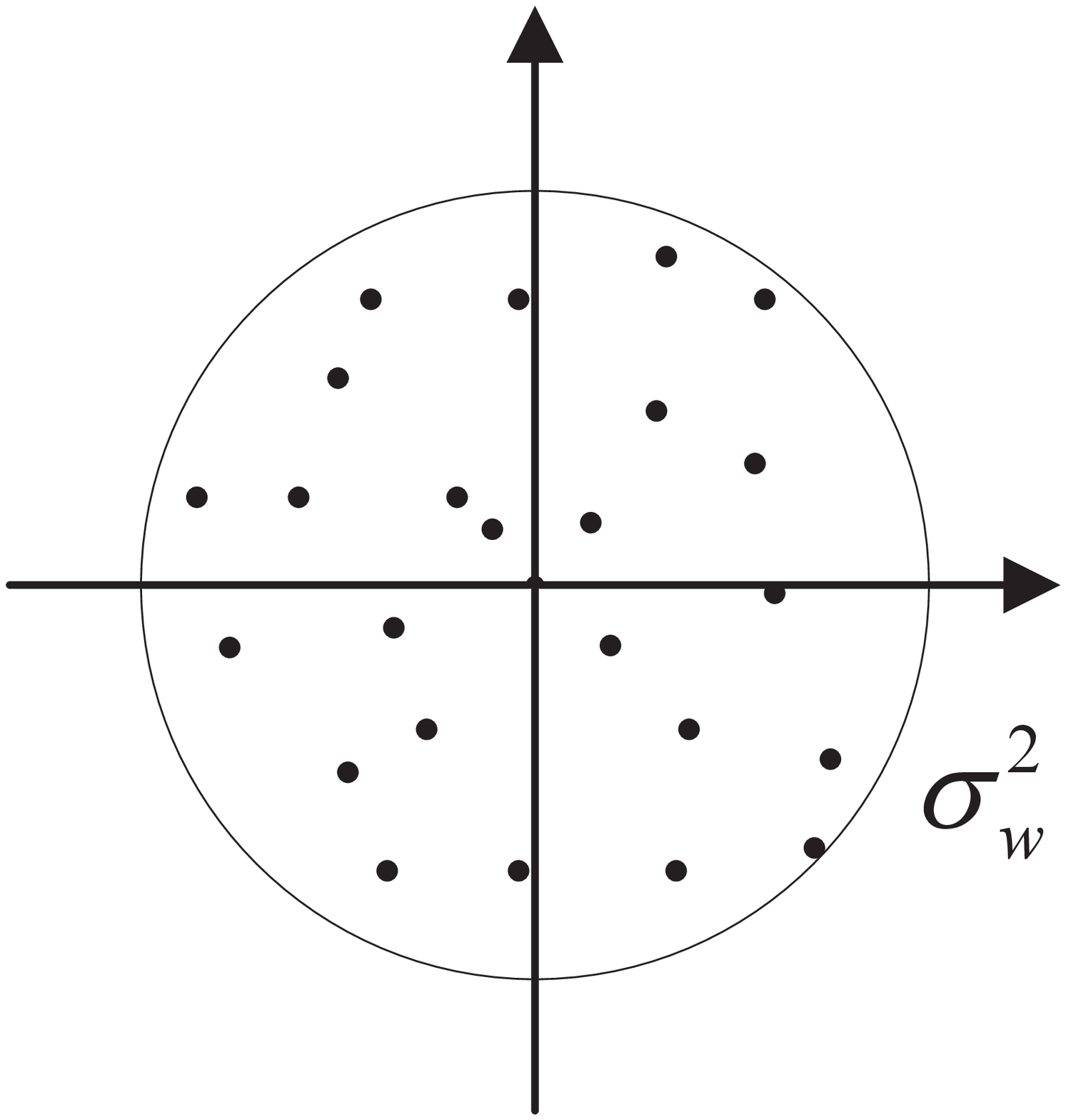}}
\caption{Noise under expectation-based model and worst-case model in two-dimensional space}
\label{noise model} 
\end{figure}

\begin{definition}[Worst-Case Robust Model \cite{vorobyov2003robust, ang2019robust}]
The expectation-based robust model refers to the stochastic property of  noise as shown in Fig. \ref{noise model} (a).
For node $j$, the entries of the uncertainty vector are assumed to be Gaussian distributed with ${\mathbb{E}} \left\{{\Delta \tilde{{\bf{w}}}_j}\right\} =  {\bf{0}}$, and ${\mathbb{E}} \left\{{\Delta \tilde{{\bf{w}}}_j}\cdot{\Delta \tilde{{\bf{w}}}_j^T}\right\} = {\sigma_{j}^2} {\bf{I}}$, $j = 1,2,...,N$, and the aggregation noise at the center is assumed to satisfy ${\mathbb{E}} \left\{{\Delta \tilde{{\bf{w}}}}\right\} =  {\bf{0}}$, and ${\mathbb{E}} \left\{{\Delta \tilde{{\bf{w}}}}\cdot{\Delta \tilde{{\bf{w}}}^T}\right\} = {\sigma^2} {\bf{I}}$.
\end{definition}

With the assumption that the aggregation noise $\Delta \tilde{{\bf{w}}}$ and the broadcast noise $\Delta \tilde{{\bf{w}}}_j$ are Gaussian, we can obtain another summed Gaussian noise as $\Delta {\bf{w}}_j$ so that the received value for node $j$ can be expressed as

\begin{equation}
\tilde{{\bf{w}}}_j = {\bf{w}} + \Delta {\bf{w}}_j, j = 1,2,...,N,\\[2mm]
\end{equation}
and $\Delta {\bf{w}}_j$ is Gaussian with ${\mathbb{E}} \left\{\Delta {\bf{w}}_j\right\} =  {\bf{0}}$, and ${\mathbb{E}} \left\{{\Delta {\bf{w}}_j}\cdot{\Delta {\bf{w}}_j^T}\right\} = {\sigma_{e_j}^2} {\bf{I}}$, $j = 1,2,...,N$, where ${\sigma_{e_j}^2} = {\sigma^2} + {\sigma_{j}^2}$.

Therefore, using the stochastic property of noise, we should focus on improving the stochastic performance for the network.
Furthermore, the optimization object in federated learning is to find the local optimal model ${{\bf{w}}}_j$ in (\ref{w_optimize_local}) and to utilize the combination method to find the global optimal  model ${{\bf{w}}}$ in (\ref{w_aggregation}).

Since the combination method is determinate, we only need to optimize the local model ${{\bf{w}}}_j$ for each node. Based on the aforementioned analysis, we formulate the robust training problem using the expectation-based model for each node as

\begin{equation}
\begin{aligned}
{\textup P_1}: \min \limits_ {{\bf{w}}}  \,\, &{\mathbb{E}}\|F_j({{\bf{w}}} + \Delta {{\bf{w}}}_j)\|^2 \\[1mm]
s.t. \,\,& \mathbb{E} \{\Delta {\bf{w}}_j\} = {\bf{0}}, j = 1,2,...,N,\\[1mm]
&\mathbb{E} \{\Delta {\bf{w}}_j \cdot \Delta {\bf{w}}_j^T\} = {\sigma _{e_j}^2} {\bf{I}}, j = 1,2,...,N,\\[2mm]
\end{aligned}
\end{equation}
where the constraints in ${\textup P_1}$ represent the stochastic characteristic of noise from imperfect estimation in wireless communication.

We aim at improving the stochastic performance for the training process. Due to the expectation calculation, the objective function is non-convex. To tackle this challenge, we consider adding the regularizer into the loss function to approximate the objective function and to represent the effect of noise. We provide the corresponding federated learning process in Section IV.

\subsection{Training Under Worst-Case Robust Model}
In  contrast to the expectation-based model, the worst-case model is a deterministic method to represent the instantaneous condition, which has fixed uncertainty sets, and to maximize the performance under the worst uncertainty \cite{shen2013worst, wang2013robust}.
Using the worst-case robust design, we can guarantee a performance level for any value of estimation realization in the uncertainty region.
It is applied to design which requires strict constraints, and is more suitable for characterizing instantaneous estimation value with errors.
The worst-case approach assumes that the actual estimation value lies in the neighborhood of the uncertainty region with a known nominal estimation value. The size of this region represents the amount of estimation value uncertainty, i.e., the bigger the region is, the more uncertainty there is.
We show the brief definition of the worst-case model as follows.

\begin{definition}[Worst-Case Robust Model \cite{vorobyov2003robust, ang2019robust}]
The worst-case robust model assumes that the  estimation lies in a known set of possible values shown as Fig. \ref{noise model} (b), which can not be exactly known.
The norm of the uncertainties vector $\Delta \tilde{{\bf{w}}}$ and $\Delta \tilde{{\bf{w}}}_j$ are bounded by the spherical region, which can be expressed as

\begin{equation}
\begin{aligned}
&{\|{\Delta \tilde{{\bf{w}}}_j}\|^2} \le {{\sigma_j^2}}, j = 1,2,...,N,\\[2mm]
&{\|{\Delta \tilde{{\bf{w}}}}\|^2} \le {{\sigma^2}}, \\[2mm]
\end{aligned}
\end{equation}
where ${{\sigma_j^2}} \ge 0$ denotes the radius of the spherical uncertainty region of the broadcast noise, while ${{\sigma^2}} \ge 0$ denotes the aggregation noise.
\end{definition}

Consider the superposition of noise, the uncertainty is expanded to the larger region with the size ${{\sigma_j^2}} + {{\sigma^2}}$. Therefore, we reformulate the received value at node $j$ as

\begin{equation}
\tilde{{\bf{w}}}_j = {\bf{w}} + \Delta {\bf{w}}_j, j = 1,2,...,N,\\[2mm]
\end{equation}
where $\Delta {\bf{w}}_j$ denotes the whole noise and satisfies ${\|{\Delta {\bf{w}}_j}\|^2} \le {{\sigma_{w_j}^2}}, j = 1,2,...,N$.

Similarly,  the optimization is to find the local optimal model ${{\bf{w}}}_j$ in (\ref{w_optimize_local}), and follows the aggregation rules in (\ref{w_aggregation}).
Therefore, we formulate the robust training problem under the worst-case model as a min-max problem for each node

\begin{equation}
\begin{aligned}
{\textup P_2}: \min \limits_{{\bf{w}}} \max \limits_{\Delta{\bf{w}}_j} \,\, &F_j({{\bf{w}}} + \Delta {{\bf{w}}}_j)\\[1mm]
s.t. \,\,& {\|{\Delta {\bf{w}}_j}\|^2} \le {{\sigma_{w_j}^2}}, j = 1,2,...,N,\\[2mm]
\end{aligned}
\end{equation}
where the constraints in ${\textup P_2}$ represent the noise lies in a spherical region with radius ${\sigma_{w_j}^2}$.

One challenge to solve the problem is that the worst condition may not be available. The other is the non-convex objective function. We settle the challenges using the sampling method and the SCA algorithm to generate a feasible descent direction for the learning process in Section V.

\section{Robust Design Using Expectation-based Model}
\label{sec:Robust Design Using Expectation-based Model}

In this section, we consider the robust design in federated learning using the expectation-based model.
We propose the corresponding RLA algorithm to represent the effects of noise for the expectation-based model so that the local optimal model can be found via optimization.

\subsection{Proposed Training Algorithm}

We first model the noise under the expectation-based noise model, which is a stochastic method to represent the random condition, as shown in ${\textup P_1}$. We aim at optimizing the average performance based on the expectation-based model. However, the random noise results in the non-convexity property and uncertainty value of the local loss function.

To solve this problem, we propose the RLA to approximate the non-convexity local loss function and utilize the distributed gradient descent to find the optimal global model.
The approximation method is inspired by previous works where training with noise was approximated via regularization to enhance the robust of neural networks \cite{graves2011practical}. We give a brief introduction in the following.

\begin{lemma}
Training with noise is equal to adding a regularizer $\Omega({{\bf{w}}})$, which can be expressed as

\begin{equation}
F(\tilde{{\bf{w}}}) \approx F({\bf{w}}) + \lambda \Omega({{\bf{w}}}),\\[2mm]
\end{equation}
where $F(\cdot)$ denotes the loss function, $\Omega(\cdot)$ is the designed function, ${\bf{w}}$ is the learning model, $\tilde{{\bf{w}}}$ represents the learning model including noise, and $\lambda$ is a constant.
\end{lemma}

\begin{IEEEproof}
Refer to \cite{goodfellow2016deep}.
\end{IEEEproof}

There are many regularization strategies in the aforementioned works \cite{bishop1995regularization, bishop1995training, graves2011practical, hochreiter1995simplifying}.
However, there is no specific regularizer that is universally  better than any others for the learning algorithm.
In other words, there is no best form of regularization.
We need to develop a specific form of $\Omega({{\bf{w}}})$ using the expectation-based model.

Motivated by this observation, we propose a new regularization term to approximate the original loss function for federated learning in the training process.
Using the expectation-based model, we intend to reduce the impact of noise for the training process.
Due to the stochastic property of noise, we aim at optimizing the average performance in ${\textup P_1}$.
We propose the corresponding training problem in the following.

\begin{proposition}[Robust Training Under Expectation-based Model]
The robust training problem under the expectation-based model in ${\textup P_1}$ for each node can be reformulated as

\begin{equation}
\label{P3}
\begin{aligned}
{\textup P_3}: \min \limits_ {{ {\bf{w}}}}  \,\, & F_j^{e}({{\bf{w}}}),\\[2mm]
\end{aligned}
\end{equation}
where $F_j^{e}({{\bf{w}}})$ denotes the new loss function for node $j$ and can be written as

\begin{equation}
\label{robust_lfunction_expectation}
F_j^{e}({{\bf{w}}})= F_j({{\bf{w}}}) + {{\sigma_e^2}}\|\nabla F_j({{\bf{w}}})\|^2.\\[2mm]
\end{equation}

\end{proposition}

\begin{IEEEproof}
Under the expectation-based model, we can obtain the objective function of ${\textup P_1}$ utilizing Taylor expansion according to the work in \cite{bishop1995training} so that the objective loss function of the optimization problem is written as

\begin{equation}
\begin{aligned}
{\mathbb{E}}\|F_j({{\bf{w}}} + {{\bf{w}}}_j)\|^2 &= {\mathbb{E}}\|F_ j({ {\bf{w}}}) + { {\bf{w}}}_j \nabla F_ j({{\bf{w}}}) + o({ {\bf{w}}})\|^2\\[1mm]
& \approx {\mathbb{E}}\|F_ j({{\bf{w}}})\|^2 + {{\sigma_e^2}}\|\nabla F_j({{\bf{w}}})\|^2.\\[2mm]
\end{aligned}
\end{equation}

The first term ${\mathbb{E}}\|F_ j({ {\bf{w}}})\|^2$ refers to the training process with perfect estimation in (\ref{w_optimize_local}), and the second term is the additional cost of the loss function in training, which is determined by noise.
Therefore, the objective loss function under the expectation-based equals adding the regularizer ${{\sigma_e^2}}\|\nabla F_j({{\bf{w}}})\|^2$.
\end{IEEEproof}

\begin{remark}
The penalty over the first-order of the loss function yields a preference for mapping $f$ that are invariant locally at the training points and drop the global model ${ {\bf{w}}}$ into the flat region.
\end{remark}

To solve the training problem in (\ref{P3}), we utilize the gradient descent algorithm to find the optimal local model ${ {\bf{w}}}$ for each node, and the details are shown as follows.

In each iteration, the local update at each node is performed based on the previous iteration and the first gradient of the proposed loss function, and the center aggregates the distributed models to find the optimal global model for the next iteration. Therefore, the update rules of the gradient descent can be written as:

\begin{subequations}
\begin{align}
&\textup{Center} : { {\bf{w}}}^{t+1} = \frac{\sum_{j=1}^N D_j {{\bf{w}}}_ j^{t+1}}{D},\\[2mm]
&\textup{Local} : {{\bf{w}}}_ j^{t+1} = { {\bf{w}}}^t - \eta \nabla F_j^{e}({ {\bf{w}}}^t), j = 1, 2, ..., N,
\end{align}
\end{subequations}\\[0.5mm]
where $\eta$ is the step size for all nodes.
The iteration is executed and it will stop if a specific condition is satisfied. This process is illustrated in Algorithm 1.

To solve the robust problem, we develop the training process by adding the regularizer to approximate the original loss function.
We transfer the stochastic and non-convex problem into a deterministic and convex problem so that we can utilize the  gradient descent method to find the optimal global model ${ {\bf{w}}}$.
The corresponding performance is shown through simulation in Section VI.

\begin{algorithm}[t]
	\renewcommand{\algorithmicrequire}{\textbf{Input:}}
	\renewcommand{\algorithmicensure}{\textbf{Output:}}
	\caption{{Distributed Gradient Descent Learning Algorithm Under Expectation-based Model} }
	\label{alg:Distributed Gradient Descent Learning Algorithm Under Expectation-based Model}
	\begin{algorithmic}[1]
		\REQUIRE ${\Delta {\bf{w}}}$, $\eta$
		\ENSURE ${{\bf{w}}}$
		\STATE number of iteration time $t$ = 0.
        \STATE Update ${{\bf{w}}}$:
		\REPEAT
        \STATE Update ${{\bf{w}}}_j^{t+1}$, $j = 1,2,...,N$ according to ${ {\bf{w}}}_ j^{t+1} = {{\bf{w}}}^t - \eta \nabla F_j^{e}({{\bf{w}}}^t)$
        \STATE Update ${ {\bf{w}}}^{t+1}$ according to the aggregated rules
		\STATE $t \leftarrow t+1$
		\UNTIL{converge}
	\end{algorithmic}
\end{algorithm}

\subsection{Convergence Analysis}
In this subsection, we derive the convergence property of the proposed design under  the expectation-based model.
To obtain the convergence rate of the proposed scheme under the expectation-based model, we first prove that the proposed federated learning is equivalent to a centralized learning, and then derive the corresponding convergence rate.

We start with the essential assumption of the loss function, which can be satisfied normally.

\begin{assumption}
We assume the following conditions for the loss function of all nodes:\\
(1) $F_ i({ {\bf{w}}})$ is $convex$,\\
(2) $F_ i({ {\bf{w}}})$ is $L-Lipschitz$, i.e. $\|F_ i({ {\bf{w}}}) - F_ i({ {\bf{w}}}^{'})\| \leq L \|{ {\bf{w}}} - { {\bf{w}}}^{'}\|$ for any ${ {\bf{w}}}$, ${ {\bf{w}}}^{'}$,\\
(3) $F_ i({ {\bf{w}}})$ is $\beta-smooth$, i.e. $\|\nabla F_ i({ {\bf{w}}}) - \nabla F_ i({ {\bf{w}}}^{'})\| \leq \beta \|{ {\bf{w}}} - { {\bf{w}}}^{'}\|$ for any ${ {\bf{w}}}$, ${ {\bf{w}}}^{'}.$
\end{assumption}

Then, we give a brief definition of  centralized learning.
\begin{definition}[Centralized learning problem under expectation-based model]
Given the proposed local loss function in (\ref{robust_lfunction_expectation}), the global loss function can be written as

\begin{equation}
\label{global}
F^{e}({ {\bf{w}}}) = \frac{{\sum_{i = 1}^N} {D_i} {F_ i^{e}({ {\bf{w}}})} }{D},\\[2mm]
\end{equation}
so that we aim at minimizing  $F^{e}({ {\bf{w}}})$ at the center by using the same whole datasets. Therefore, the centralized learning problem is to find the optimal global model as

\begin{equation}
\begin{aligned}
{\textup P_4}: \min \limits_ {{ {\bf{w}}}}  \,\, & F^{e}({ {\bf{w}}}).\\[2mm]
\end{aligned}
\end{equation}

\end{definition}

The optimization can be easily solved by using the gradient descent, and the center completes the iteration until the specific condition is met.
We derive that the proposed federated learning is equivalent to the centralized learning problem under the expectation-based model as follows.

\begin{lemma}
Given ${\textup P_1}$ and  ${\textup P_4}$ under the expectation-based model, the proposed federated learning is equal to the centralized learning for each iteration $t$, $t =  0, 1, 2, ...$, which can be written as

\begin{equation}
{ {\bf{w}}}^{t+1} = { {\bf{w}}}^t - \eta \nabla F^{e}({ {\bf{w}}}^t).\\[2mm]
\end{equation}
\end{lemma}

\begin{IEEEproof}
Considering the global aggregation, we can obtain that
\begin{equation}
\begin{aligned}
{ {\bf{w}}}^{t+1} &= \frac{\sum_{j=1}^N D_j { {\bf{w}}}_ j^{t+1}}{D}\\[1mm]
&= \frac{\sum_{j=1}^N D_j({ {\bf{w}}}^t - \eta F^{e}_j({ {\bf{w}}}^t))}{D}\\[1mm]
&= \frac{\sum_{j=1}^N D_j { {\bf{w}}}^t}{D} - \eta \frac{\sum_{j=1}^N D_j F^{e}_j({ {\bf{w}}}^t)}{D}\\[1mm]
&= { {\bf{w}}}^t - \eta \nabla F^{e}({ {\bf{w}}}^t).\\[2mm]
\end{aligned}
\end{equation}
\end{IEEEproof}

To prove the convergence of the proposed distributed learning, we only need to derive that the equivalent centralized learning is convergent.


\begin{lemma}
Given the original loss function under Assumption 1, there exist constants $\eta$ and $\beta$ so that the loss function $f_j({{\bf{w}}}),  \,\,j=1, 2, ..., N$ satisfies that

\begin{equation}
\label{centralized_cv}
F_i({ {\bf{w}}}^t) - F_i({ {\bf{w}}}^*) \leq \|{ {\bf{w}}}^0 - { {\bf{w}}}^*\|^2 \cdot {\frac{1}{\eta\left(1 - {\frac{\beta\eta}{2}}\right)}} \cdot {\frac{1}{t}},\\[2mm]
\end{equation}
where ${ {\bf{w}}}^0$ is the initialization point of ${ {\bf{w}}}$.
\end{lemma}
\begin{IEEEproof}
Refer to \cite{chong2013introduction}.
\end{IEEEproof}

\begin{lemma}
$F({ {\bf{w}}})$ is $convex$, $L-Lipschitz$ and $\beta-smooth$.
\end{lemma}

\begin{IEEEproof}
We can obtain that  $F({ {\bf{w}}})$ is the linear combination of $F_i({ {\bf{w}}})$ via (\ref{global}). Straightforwardly from the convexity property, this lemma holds.
\end{IEEEproof}

\begin{proposition}[Convergence Under Worst-case Model]
Algorithm 1 yields the following convergence property for the optimization of the global loss function under the expectation-based model

\begin{equation}
\label{conv_exp}
F^{e}({ {\bf{w}}}^t) - F^{e}({ {\bf{w}}}^*) \leq \|{ {\bf{w}}}^0 - { {\bf{w}}}^*\|^2 \cdot {\frac{1}{\eta\left(1 - {\frac{(1 + \lambda{{\sigma_e^2}}) \beta\eta}{2}}\right)}} \cdot {\frac{1}{t}},\\[2mm]
\end{equation}
where ${ {\bf{w}}}^0$ is the initialization point of ${ {\bf{w}}}$.
It means the convergence rate is ${\mathcal{O}({1}/{t})}$.

\end{proposition}

\begin{IEEEproof}
The proposed loss function of the node $j$ is

\begin{equation}
F_j^{e}({ {\bf{w}}}) = F_j({ {\bf{w}}}) + {{\sigma_e^2}}\|\nabla F_j({ {\bf{w}}})\|^2.\\[2mm]
\end{equation}

Taking the derivation of it, we can obtain

\begin{equation}
\begin{aligned}
\nabla F_j^{e}({ {\bf{w}}}) & = \nabla F_j({ {\bf{w}}}) + {{\sigma_e^2}} \nabla {\rm{tr}} ({\nabla F_j({ {\bf{w}}}) \nabla F_j({ {\bf{w}}})^T})\\[1mm]
&= \nabla F_j({ {\bf{w}}}) + {{\sigma_e^2}} \nabla F_j({ {\bf{w}}})\\[1mm]
& = (1 + {{\sigma_e^2}}) \nabla F_j({ {\bf{w}}}).\\[2mm]
\end{aligned}
\end{equation}

Following the Lemma 4, we can obtain that the loss function $F_j^{e}({ {\bf{w}}})$ of the node $j$ is $\beta-smooth$ with $\tilde{\beta} = (1 + {{\sigma_e^2}}) \beta$.
Therefore, $F_j^{e}({ {\bf{w}}})$ satisfies

\begin{equation}
\label{conv_exp_local}
F_j^{e}({ {\bf{w}}}) - F_j^{e}({ {\bf{w}}}^*) \leq \|{ {\bf{w}}}^0 - { {\bf{w}}}^*\|^2 \cdot {\frac{1}{\eta\left(1 - {\frac{(1 + {{\sigma_e^2}}) \beta\eta}{2}}\right)}} \cdot {\frac{1}{t}}.\\[2mm]
\end{equation}

Furthermore, we can develop the conclusion that $F^{e}({ {\bf{w}}})$ is $\beta-smooth$ to satisfy (\ref{conv_exp}). The optimization of the global loss function converges at ${\mathcal {O}}(1/t)$.

\end{IEEEproof}

\begin{remark}
The proposed robust design under the expectation-based model converges at ${\mathcal {O}}(1/t)$.
The convergence property as (\ref{conv_exp}) is reduced to the one in (\ref{centralized_cv}) as $\sigma_e^2 = 0$, i.e., it is equivalent to the convergence property that is training  without noise.
The convergence rate will decrease with the increase in $\sigma_e^2$ and the proposed design cannot converge when $\left(1 - {{(1 + {{\sigma_e^2}}) \beta\eta}/{2}}\right) \leq 0$ specifically.
The comparison between the proposed design and the centralized training is simulated specifically in Section VI.
\end{remark}


\section{Robust Design Using Worst-case Model}
\label{sec:Robust Design Using Worst-case Model}
In this section, we solve the optimization problem using the worst-case model. To solve the uncertainty of noise and the non-convexity problem, we utilize the sampling-based SCA method to represent noise and approximate the objective loss function. We then propose the training process for the robust federated learning and  finally derive the convergence property of the proposed design.

\subsection{Proposed Training Algorithm}
The training process is proposed to solve the learning problem under the worst-case model. We utilize the sampling-based SCA method to approximate the original objective function, and develop the corresponding updating rules.

The feasible sets of both the local model and the noise are convex sets, and there always exists a saddle point.
However, the unavailability of noise results in that the finding of the global minimum point is, in general, NP hard.
Therefore, the objective problem faces the main issues:  i) the impossibility to estimate accurate value of noise of the worst condition; ii) the non-convexity of the objective functions leading to unavailable optimization.

Considering the uncertainty of noise, it is often possible to obtain a sample of the random noise, either from past data or from computer simulation as shown in \cite{kleywegt2002sample}. Consequently, one may consider an approximate solution to the problem based on sampling, known as the sample average approximation (SAA) method, and we give a brief introduction as follows.

\begin{lemma}
The SAA method is to find the optimal $x$ for the stochastic objective in the optimization problem as,

\begin{equation}
x^* = \min {\mathbb{E}}[f(x;\xi)],\\[2mm]
\end{equation}
where $f(x;\xi)$ is a given function and affected by the random vector $\xi$ which follows the distribution $V$. However, the distribution $V$ is unknown, and only sample values of the random vector $\xi$ are available. To solve this problem, the SAA approach approximates the problem by solving

\begin{equation}
\hat{x}^* = \min \frac{1}{N} \sum_{j=1}^N f(x;\xi^j),\\[2mm]
\end{equation}
where $\xi^j$ is the random sample of the random vector $\xi$, and the collection of $N$ realizations satisfies independent and identically distributed.
\end{lemma}

\begin{IEEEproof}
Refer to \cite{kleywegt2002sample}.
\end{IEEEproof}

Motivated by this method, we consider sampling  noise ${\Delta {\bf{w}}}_j$ in the objective function $F_j({{ {\bf{w}}}} +  {{\Delta {\bf{w}}}}_j)$, and can easily obtain that the worst condition of  noise occurs on the boundary.
Based on the above consideration, we propose the sampling-based method. At each iteration $t$ of each node, a new realization of the noise ${\Delta {\bf{w}}}_j^t$ is obtained and the optimization of the objective functions is updated via the loss function as follows,

\begin{equation}
F_j({{ {\bf{w}}}} +  {{\Delta {\bf{w}}}}_j) = F_j({ {\bf{w}}} + {\Delta {\bf{w}}}_j^t), t = 1, 2, ....\\[2mm]
\end{equation}
where ${\Delta {\bf{w}}}_j^t$ satisfies $\|{\Delta {\bf{w}}}_j^t\|^2 = \sigma_w^2$.

It provides a simple way to approach the objective function under the perfect estimation, but the non-convexity of the objective function is still not resolved.
To tackle this challenge, we utilize the SCA scheme to maintain the convexity of the objective functions.

\begin{lemma}
The SCA algorithm is proposed to approximate an arbitrarily function $f(x)$ by expansion around $x^t$ which is a definite point in the feasible set. It can be simply written as

\begin{equation}
f(x) \approx \tilde{f}(x, x^t) = \rho^t f(x) + (1 - \rho^t) \langle{x - x^t},{ g(x^t)}\rangle,\\[2mm]
\end{equation}
where $\rho^t\in(0,1]$ is a sequence, and $g(x^t)$ is the weight average of the first gradient and can be expressed as

\begin{equation}
g(x^t) = (1 - \rho^t) g(x^{t-1}) + \rho^t \nabla f(x^t).
\end{equation}
\end{lemma}

\begin{IEEEproof}
Refer to \cite{yang2016parallel}.
\end{IEEEproof}

With the consideration of SAA and SCA methods, we propose the sampling-based SCA algorithm to solve the robust training problem under the worst-case model of ${\textup P_2}$ in the following.

\begin{proposition}[Robust Training Under Worst-case Model]
For the robust training problem under the worst-case model in ${\textup P_2}$,the optimization problem of each node can be reformulated as

\begin{equation}
\begin{aligned}
{\textup P_5}: \min \limits_ {{ {\bf{w}}}}  \,\, & F_j^{w} ({ {\bf{w}}}; { {\bf{w}}}^t,  {\Delta {\bf{w}}}_j^t)\\[2mm]
\end{aligned}
\end{equation}
where $ {\Delta {\bf{w}}}_j^t$ is a sequence by sampling the noise $ {\Delta {\bf{w}}}_j$ satisfying that $\|{\Delta {\bf{w}}}_j^t\|^2 = \sigma_w^2$, $F_j^{w} ({ {\bf{w}}}; { {\bf{w}}}^t,  {\Delta {\bf{w}}}_j^t)$ is denoted as the loss function for the node $j$, and expressed as

\begin{equation}
\label{lfunction_worstcase}
\begin{aligned}
F_j^{w} ({ {\bf{w}}}; { {\bf{w}}}^t,  {\Delta {\bf{w}}}_j^t) = &\rho^t {F_j}({ {\bf{w}}} +  {\Delta {\bf{w}}}_j^t) + \lambda \|{ {\bf{w}}}-{ {\bf{w}}}^t\|^2  \\
& + (1 - \rho^t)\langle{{ {\bf{w}}} - { {\bf{w}}}^t},{G_j^{t-1}}\rangle,\\[2mm]
\end{aligned}
\end{equation}
and $G_j^t$ is an accumulation vector updated recursively according to

\begin{equation}
\label{gradient_worstcase}
G_j^t = (1 - {\rho^t}){G_j^{t-1}} + {\rho^t} \nabla_{{ {\bf{w}}}_j} {F_j}({ {\bf{w}}} +  {\Delta {\bf{w}}}_j^t),\\[2mm]
\end{equation}
with $\rho^t\in(0,1]$ being a sequence to be properly chosen $(\rho^0 =1)$, $t = 0, 1, ...$.

\end{proposition}

\begin{IEEEproof}
As the efficient solutions of the SCA algorithm, the objective function $F_j({ {\bf{w}}}, {\Delta {\bf{w}}}_j^t)$ at the iteration $t$ is determined by the latest updated model ${\Delta {\bf{w}}}_j^t$ and defined as $F_j^{w}({ {\bf{w}}}; { {\bf{w}}}^t, {\Delta {\bf{w}}}_j^t)$, which is consist of the original function ${F_j}({ {\bf{w}}} + {\Delta {\bf{w}}}_j^t)$, and the first gradient ${ \Omega_1({ {\bf{w}}}) = \langle{{ {\bf{w}}} - { {\bf{w}}}^t},{G_j^{t-1}}\rangle}$.
We develop the objective function as follows,

\begin{equation}
F_j^{w} ({ {\bf{w}}}; { {\bf{w}}}^t,  {\Delta {\bf{w}}}_j^t) = \rho^t {F_j}({ {\bf{w}}} +  {\Delta {\bf{w}}}_j^t) + (1 - \rho^t)\langle{{ {\bf{w}}} - { {\bf{w}}}^t},{G_j^{t-1}}\rangle,\\[2mm]
\end{equation}
and $G_j^t$ is an accumulation vector updated recursively according to

\begin{equation}
\label{gradient_worstcase}
G_j^t = (1 - {\rho^t}){G_j^{t-1}} + {\rho^t} \nabla_{ {\bf{w}}} {F_j}({ {\bf{w}}} +  {\Delta {\bf{w}}}_j^t),\\[2mm]
\end{equation}
with $\rho^t\in(0,1]$ being a sequence to be properly chosen $(\rho^0 =1)$ at iterations $t=0,1,...$ respectively.


Notice that the expansion is established only when ${ {\bf{w}}}$ is close to ${ {\bf{w}}}^t$. We add a regularizer as the cost of shrinking the gap between ${ {\bf{w}}}$ and ${ {\bf{w}}}^t$ as:

\begin{equation}
\Omega_2 ({ {\bf{w}}})=  \|{ {\bf{w}}} - { {\bf{w}}}^t\|^2.\\[2mm]
\end{equation}
Therefore, we propose the local loss function as in ${\textup P_4}$.
\end{IEEEproof}


\begin{remark}
Generally speaking, each node minimizes the sample approximation of the original unstable function.
The first term in ({\ref{lfunction_worstcase}}) refers to the sample objective function.
The second term refers to the cost which controls the pace for each iteration.
The vector $G_j^t$ in the last term represents the incremental estimate of the unknown $\nabla_{ {\bf{w}}} {F_j}({ {\bf{w}}} +  {\Delta {\bf{w}}}_j^t)$ by samples collection over the iterations.
When the parameter ${\rho^t}$ is properly chosen, and the estimation accuracy increases as $t$ increases.
\end{remark}

Due to the involving of the past optimized model ${ {\bf{w}}}^t$, we consider utilizing the conditional gradient descent method for each node.
Similarly, we aggregate the local update at the center and broadcast the new global model for next iteration.
The aggregated model ${ {\bf{w}}}^{t+1}$ should be broadcasted to all nodes and it is used to complete the next iteration until it meets the specific condition.
Given ${ {\bf{w}}}_j^{w} ({ {\bf{w}}}^t,  {\Delta {\bf{w}}}_j^t) = \arg\min F_j^{w} ({ {\bf{w}}}; { {\bf{w}}}^t,  {\Delta {\bf{w}}}_j^t)$, the iteration rule is briefly written as follows.

\begin{subequations}
\begin{align}
&\textup{Center} : { {\bf{w}}}^{t+1} = \frac{\sum_{j=1}^N D_j { {\bf{w}}}_ j^{t+1}}{D},\\[2mm]
&\textup{Local} : { {\bf{w}}}_j^{t+1} = { {\bf{w}}}^t + \gamma^{t+1}\left({ {\bf{w}}}_j^{w} ({ {\bf{w}}}^t,  {\Delta {\bf{w}}}_j^t) - { {\bf{w}}}^t\right),
\end{align}
\end{subequations}\\[0.5mm]
where $\gamma^t \in(0,1]$, $t = 0, 1, 2,...$.
The iteration follows the process illustrated in Algorithm 2.

\begin{algorithm}[t]
	\renewcommand{\algorithmicrequire}{\textbf{Input:}}
	\renewcommand{\algorithmicensure}{\textbf{Output:}}
	\caption{{Distributed Gradient Descent Learning Algorithm Under Worst-case Model} }
	\label{alg:Distributed Gradient Descent Learning Algorithm Under Worst-case Model}
	\begin{algorithmic}[1]
		\REQUIRE ${\Delta {\bf{w}}}$, $\gamma$
		\ENSURE ${ {\bf{w}}}$
		\STATE number of iteration time $t$ = 0.
        \STATE Update ${ {\bf{w}}}$:
		\REPEAT
        \STATE Update ${ {\bf{w}}}_j^{t+1}$, $j = 1,2,...,N$ according to ${ {\bf{w}}}_j^{t+1} = { {\bf{w}}}^t + \gamma^{t+1}\left({ {\bf{w}}}_j^{w} ({ {\bf{w}}}^t,  {\Delta {\bf{w}}}_j^t) - { {\bf{w}}}^t\right)$
        \STATE Update ${ {\bf{w}}}^{t+1}$ according to the aggregated rules
		\STATE $t \leftarrow t+1$
		\UNTIL{converge}
	\end{algorithmic}
\end{algorithm}

We develop the training process by utilizing the sampling-based SCA algorithm to approximate the training objective function for each node. With the iteration between the conditional gradient descent and the aggregation step, we can obtain the optimal global model ${ {\bf{w}}}$. The corresponding performance is shown through simulations in Section VI.

\subsection{Convergence Analysis}
To obtain the convergence rate of the proposed scheme under the worst-case model, we similarly prove that the proposed federated learning is equal to the centralized learning, and then derive the corresponding convergence rate.

Without loss of generality, we first give some assumptions before the further analysis.
\begin{assumption}
We assume the following conditions for the loss function of all nodes\\
(1) $\tilde{F}_ i({ {\bf{w}}}, {\Delta {\bf{w}}})$ is convex,\\
(2) $\tilde{F}_ i({ {\bf{w}}}, {\Delta {\bf{w}}})$ is $L-Lipschitz$, i.e., $\|\tilde{F}_ i({ {\bf{w}}} , {\Delta {\bf{w}}}) - \tilde{F}_ i({ {\bf{w}}}^{'}, {\Delta {\bf{w}}}^{'})\| \leq L \|{ {\bf{w}}} - { {\bf{w}}} ^{'}\|$ for any ${ {\bf{w}}} $, ${ {\bf{w}}} ^{'}$ and $ {\Delta {\bf{w}}}$,\\
(3) $\tilde{F}_ i(\tilde{{ {\bf{w}}}}, {\Delta {\bf{w}}})$ is $\beta-smooth$, i.e., $\|\nabla\tilde{F}_ i({ {\bf{w}}} , {\Delta {\bf{w}}}) - \nabla\tilde{F}_ i({ {\bf{w}}} ^{'}, {\Delta {\bf{w}}})\| \leq \beta \|{ {\bf{w}}} - { {\bf{w}}} ^{'}\|$ for any ${ {\bf{w}}} $, ${ {\bf{w}}} ^{'}$ and $ {\Delta {\bf{w}}}$.
\end{assumption}

We first develop a brief introduction of the optimization problem in centralized learning under the worst-case model.
\begin{definition}[Centralized learning problem under worst-case model]
Given the local loss function in (\ref{lfunction_worstcase}), we can obtain that the global loss function in iteration $t$ is

\begin{equation}
\label{lfunction_global_worst_case}
F^{w}({ {\bf{w}}}; { {\bf{w}}}^t, {\Delta {\bf{w}}}^t) = \frac{{\sum_{i = 1}^N} {D_i} {F_ i^{w}({ {\bf{w}}}; { {\bf{w}}}^t, {\Delta {\bf{w}}}^t)} }{D},\\[2mm]
\end{equation}
where ${ {\bf{w}}}^t$ is the global model in last iteration $t-1$, and ${\Delta {\bf{w}}}^t$ denotes the sampled noise in last iteration $t-1$, which satisfies $\|{\Delta {\bf{w}}}^t\|^2 \leq \sigma_w^2$.

Due to the fact that we aim at minimizing the global loss function, the centralized learning problem is to find the optimal global model ${\bf{w}}$ in iteration $t$, i.e.,

\begin{equation}
\begin{aligned}
{\textup P_6}: \min \limits_ {{ {\bf{w}}}}  \,\, & F^{w}({ {\bf{w}}}; { {\bf{w}}}^t, {\Delta {\bf{w}}}^t).\\[2mm]
\end{aligned}
\end{equation}

The problem can be solved by the SCA algorithm, and the center completes the iteration until it meets the specific condition.

\end{definition}

In the following, we first prove that the federated learning is equivalent to the centralized learning under the worst-case model. Secondly, we show that the centralized learning under the worst-case model is convergent.

\begin{lemma}
Given the problem under Assumption 2, suppose that $\tau > 0$ and step size $\gamma^t$ and $\rho^t$ are chosen as $\gamma_i^t = \gamma^t = \frac{1}{t^\alpha}$ and $\rho_i^t = \rho^t = \frac{1}{t^\beta}$, $0.5 < \beta < \alpha < 1$, $i = 1, 2, ..., N$ so that the distributed learning equals the centralized learning at iteration $t$, which is expressed as

\begin{equation}
\label{centerlized_worst_case}
{ {\bf{w}}}^{w}({ {\bf{w}}}^t, {\Delta {\bf{w}}}^t) = \arg\min  F^{w} ({ {\bf{w}}}; { {\bf{w}}}^t, {\Delta {\bf{w}}}^t),\\[2mm]
\end{equation}
and the global model aggregation obeys the updating rules as

\begin{equation}
\label{w_global_worst_case}
{ {\bf{w}}}^{t+1} = { {\bf{w}}}^t + \gamma^{t+1}({ {\bf{w}}}^{w}({ {\bf{w}}}^t, {\Delta {\bf{w}}}^t)- { {\bf{w}}}^t).\\[2mm]
\end{equation}

\end{lemma}

\begin{IEEEproof}
For any iteration $t+1$, ${ {\bf{w}}}^{t+1}$ satisfies
\begin{equation}
\begin{aligned}
{ {\bf{w}}}^{t+1} = & \frac{\sum_{j=1}^N D_j { {\bf{w}}}_ j^{t+1}}{D}\\[1mm]
= &\frac{\sum_{j=1}^N D_j \left[{ {\bf{w}}}^t + \gamma^{t+1}\left({ {\bf{w}}}_j^{w} ({ {\bf{w}}}^t,  {\Delta {\bf{w}}}_j^t) - { {\bf{w}}}^t\right)\right]}{D}\\[1mm]
= & \frac{\sum_{j=1}^N D_j { {\bf{w}}}^t}{D} + \gamma^{t+1}{\frac{\sum_{j=1}^N D_j { {\bf{w}}}_j^{w} ({ {\bf{w}}}^t, {\Delta {\bf{w}}}^t)}{D}} \\[1mm]
&  - \gamma^{t+1}{\frac{\sum_{j=1}^N D_j { {\bf{w}}}_j^t}{D}}\\[1mm]
= & { {\bf{w}}}^t + \gamma^{t+1}({ {\bf{w}}}^{w}({ {\bf{w}}}^t, {\Delta {\bf{w}}}^t)- { {\bf{w}}}^t).\\[2mm]
\end{aligned}
\end{equation}
\end{IEEEproof}

To prove the convergence of the distributed learning, we only need to prove that the equivalent centralized learning is convergent.

\begin{lemma}
Given the problem under Assumption 2, we can achieve that the global loss function $F^{w}({ {\bf{w}}}; { {\bf{w}}}^t, {\Delta {\bf{w}}}^t)$ satisfies Assumption 2.
\end{lemma}

\begin{IEEEproof}
According to the aggregation rules, the global loss function $F^{w}({ {\bf{w}}}; { {\bf{w}}}^t, {\Delta {\bf{w}}}^t)$ is written in (\ref{lfunction_global_worst_case}), which is the linear combination of the local loss function $F_j^{w}({ {\bf{w}}}; { {\bf{w}}}^t, {\Delta {\bf{w}}}^t)$. Straightforwardly from the convexity property, we can derive the conclusion.
\end{IEEEproof}

\begin{proposition}[Convergence Under Worst-case Model]
Given problem under Assumption 2, suppose that $\tau > 0$ and step size $\gamma^t$ and $\rho^t$ are chosen as $\gamma^t = \frac{1}{t^\alpha}$ and $\rho^t = \frac{1}{t^\beta}$, $0.5 < \beta < \alpha < 1$ for the centralized learning. Let $\left\{{ {\bf{w}}}^t\right\}$ be the sequence generated by algorithm, $F^{w}({ {\bf{w}}}; { {\bf{w}}}^t, {\Delta {\bf{w}}}^t)$ be $F^{w}({ {\bf{w}}})$ and ${ {\bf{w}}}^{w}({ {\bf{w}}}^t, {\Delta {\bf{w}}}^t)$ be ${ {\bf{w}}}^{{w},t}$. The global loss function $F^{w}({ {\bf{w}}})$ converges at ${\mathcal {O}}(\gamma^t)$ so that there exists a constant $M$ satisfying

\begin{equation}
\label{conv_worst_case}
F^{w}({ {\bf{w}}}^t) - F^{w}({ {\bf{w}}}^*) \leq M \gamma^{t}.\\[2mm]
\end{equation}
\end{proposition}

\begin{IEEEproof}
Firstly, we can obtain that $G^t = {{\sum_{i = 1}^N} {D_i} {G_i^t} }/{D}$, and $\tilde{F}^t = {{\sum_{i = 1}^N} {D_i} {\tilde{F}_i^t} }/{D}$ via the updating rules.
Furthermore, according to lemma, we have that $\tilde{F}^t$ also satisfies the Assumption 2.
Invoking the first-order optimality conditions of $F^{w}({ {\bf{w}}})$, we have
\begin{equation}
\begin{aligned}
&\rho^t \left\langle{ {\bf{w}}}^t - { {\bf{w}}}^{{w},t}, \nabla {\tilde{F}}({ {\bf{w}}}^{{w},t},  {\Delta {\bf{w}}}_j^t)\right\rangle \\[1mm]
&+ \lambda \|{ {\bf{w}}}^t - { {\bf{w}}}^{{w},t}\|^2 + (1 - \rho^t) \left\langle{ {\bf{w}}}^t - { {\bf{w}}}^{{w},t}, G_j^t\right\rangle\\[1mm]
= & \rho^t \left\langle{ {\bf{w}}}^t - { {\bf{w}}}^{{w},t}, \nabla {\tilde{F}}({ {\bf{w}}}^{{w},t},  {\Delta {\bf{w}}}^t) - \nabla {\tilde{F}}({ {\bf{w}}}^t,  {\Delta {\bf{w}}}^t)\right\rangle \\[1mm]
&+ \left\langle{ {\bf{w}}}^t - { {\bf{w}}}^{{w},t}, G^t\right\rangle + \lambda \|{ {\bf{w}}}^t - { {\bf{w}}}^{{w},t}\|^2 \geq 0\\[1mm]
\end{aligned}
\end{equation}

Considering the convexity of the ${\tilde{F}}(\cdot,  {\Delta {\bf{w}}}^t)$ , we can obtain that

\begin{equation}
\left\langle{ {\bf{w}}}^t - { {\bf{w}}}^{{w},t}, G^t\right\rangle \leq - \lambda \|{ {\bf{w}}}^t - { {\bf{w}}}^{{w},t}\|^2.\\[2mm]
\end{equation}

Given $F^{w}({ {\bf{w}}})$ under the Assumption 2, there will exist a constant $L$ so that
\begin{equation}
\begin{aligned}
F^{w}({ {\bf{w}}}^{t+1})  \leq \,\,& F^{w}({ {\bf{w}}}^t) + \gamma^{t+1} \left\langle{ {\bf{w}}}^t - { {\bf{w}}}^{{w},t}, \nabla F^{w}({ {\bf{w}}}^t)\right\rangle \\[1mm]
& +  L (\gamma^{t+1})^2 \|{ {\bf{w}}}^t - \hat{{ {\bf{w}}}}^t\|^2\\[1mm]
= \,\, & F^{w}({ {\bf{w}}}^t) + L (\gamma^{t+1})^2 \|{ {\bf{w}}}^t - { {\bf{w}}}^{{w},t}\|^2  \\[1mm]
& +  \gamma^{t+1} \left\langle{ {\bf{w}}}^t - { {\bf{w}}}^{{w},t}, \nabla F^{w}({ {\bf{w}}}^t) - G^t + G^t \right\rangle\\[1mm]
\leq \,\, & F^{w}({ {\bf{w}}}^t) - \gamma^{t+1} (\lambda - L \gamma^{t+1})\|{ {\bf{w}}}^t - { {\bf{w}}}^{{w},t}\|^2 \\[1mm]
& +  \gamma^{t+1}\|{ {\bf{w}}}^t - { {\bf{w}}}^{{w},t}\| \|\nabla F^{w}({ {\bf{w}}}^t) - G^t\|\\[2mm]
\end{aligned}
\end{equation}

Suppose that $\lim_{t \to +\infty}\|{ {\bf{w}}}^t - \hat{{ {\bf{w}}}}^t\| \geq W \geq 0$, so that we can derive that
\begin{equation}
\begin{aligned}
F^{w}({ {\bf{w}}}^{t+1}) \leq \,\, &F^{w}({ {\bf{w}}}^t) - \gamma^{t+1} (\lambda - L \gamma^{t+1}) W^2 \\[1mm]
& + \gamma^{t+1} \|\nabla F^{w}({ {\bf{w}}}^t) - G^t\| W.\\[2mm]
\end{aligned}
\end{equation}

We focus on a realization that $\lim_{t \to +\infty}\|\nabla F^{w}({ {\bf{w}}}^t) - G^t\| = 0$. Therefore, there exists a $t_0$ sufficiently large so that for $\forall t > t_0$

\begin{equation}
\lambda - L \gamma^{t+1} - \frac{1}{W}\|\nabla F^{w}({ {\bf{w}}}^t) - G^t\| \geq \tilde{\lambda} > 0.\\[2mm]
\end{equation}
Therefore, the global loss function $F^{w}({ {\bf{w}}})$ follows

\begin{equation}
F^{w}({ {\bf{w}}}^{t+1}) - F^{w}({ {\bf{w}}}^t) \leq - \tilde{\lambda} W^2 \gamma^{t+1}.\\[2mm]
\end{equation}

We show next that the gap between the $F^{w}({ {\bf{w}}}^t)$ and the optimal $F^{w}({ {\bf{w}}}^{t^*})$ is

\begin{equation}
\begin{aligned}
F^{w}({ {\bf{w}}}^t) - F^{w}({ {\bf{w}}}^*) = & \left[F^{w}({ {\bf{w}}}^t) - F^{w}({ {\bf{w}}}^{t_0})\right] \\[1mm]
& + \left[F^{w}({ {\bf{w}}}^{t_0}) - F^{w}({ {\bf{w}}}^{t^*})\right]\\[1mm]
\leq & \tilde{\lambda} W^2\left(- \sum_{m = t_0}^t \gamma^m - \sum_{m = t^*}^{t_0} \gamma^m\right)\\[1mm]
= &\tilde{\lambda} W^2\left(-\frac{\gamma^{t_0}(1 - \gamma^t)}{1-\gamma}\right) \\[1mm]
& - \tilde{\lambda} W^2 \left( \frac{\gamma^{t^*}(1 - \gamma^{t_0})}{1-\gamma}\right)\\[1mm]
\leq & \tilde{\lambda} W^2 \frac{1}{1-\gamma} \gamma^t.\\[2mm]
\end{aligned}
\end{equation}

Let the constant $M$ satisfy $M = \tilde{\lambda} W^2 / {(1-\gamma)}$, and we obtain the convergence rate expression in (\ref{conv_worst_case}).

\end{IEEEproof}

\begin{remark}
The proposed robust design under the expectation-based model converges at ${\mathcal {O}}(\gamma^t)$. The centralized training process converges at ${\mathcal {O}}(1/t)$, which utilizes the gradient descent under perfect estimation.
Compared with the centralized training, the proposed design converges at a higher speed when the iteration time $t$ increases.
The comparison between the proposed design and the centralized training is simulated specifically in Section VI.
\end{remark}

\section{Simulation Results and Discussions}

\begin{figure}[!t]
\centering
\subfigure[The accuracy performance versus iterative times under expectation-based model.]{
\begin{minipage}[t]{0.45\textwidth}
\centering
\includegraphics[width=\textwidth]{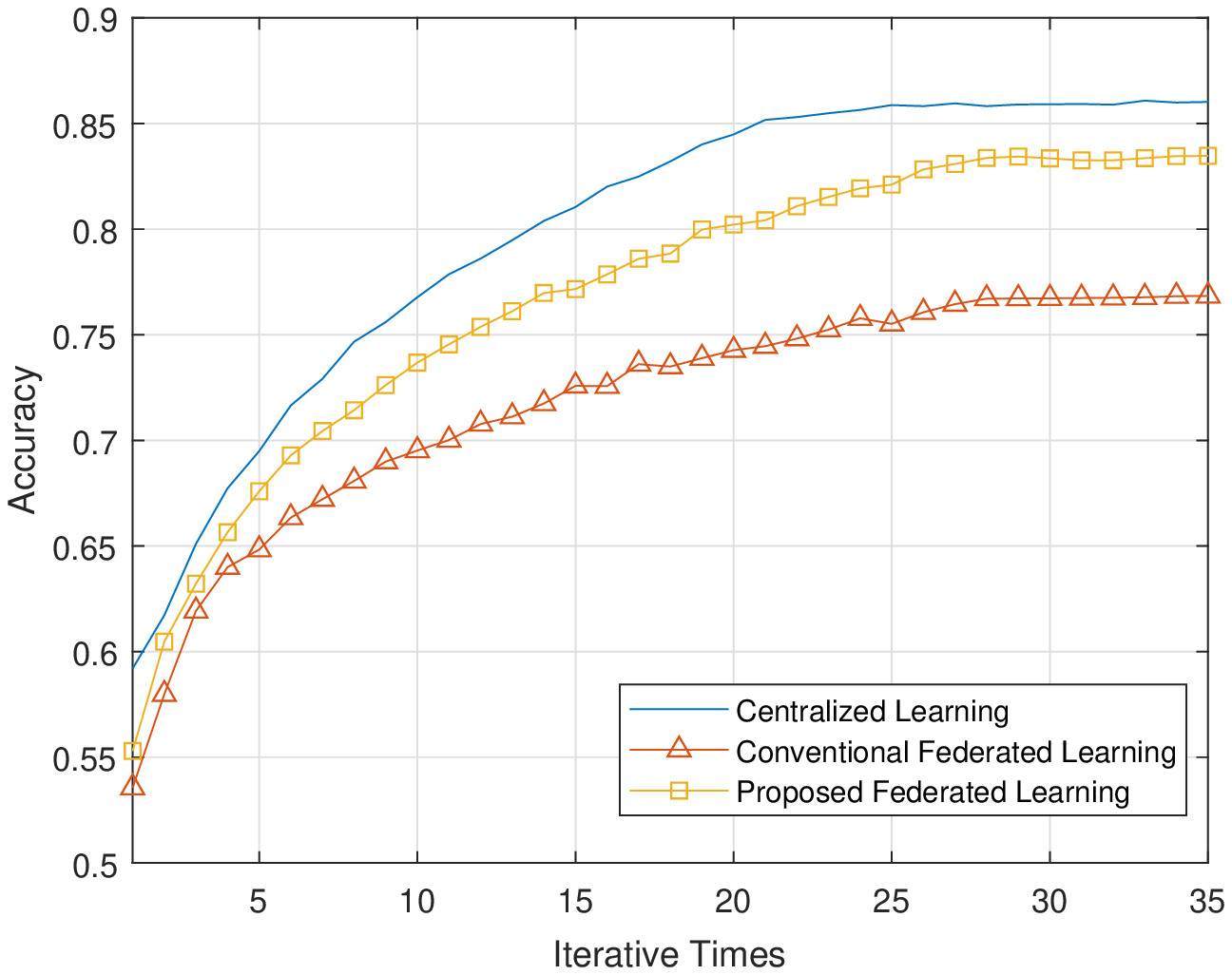}
\end{minipage}%
}%

\subfigure[The loss function performance versus iterative times under expectation-based model.]{
\begin{minipage}[t]{0.45\textwidth}
\centering
\includegraphics[width=\textwidth]{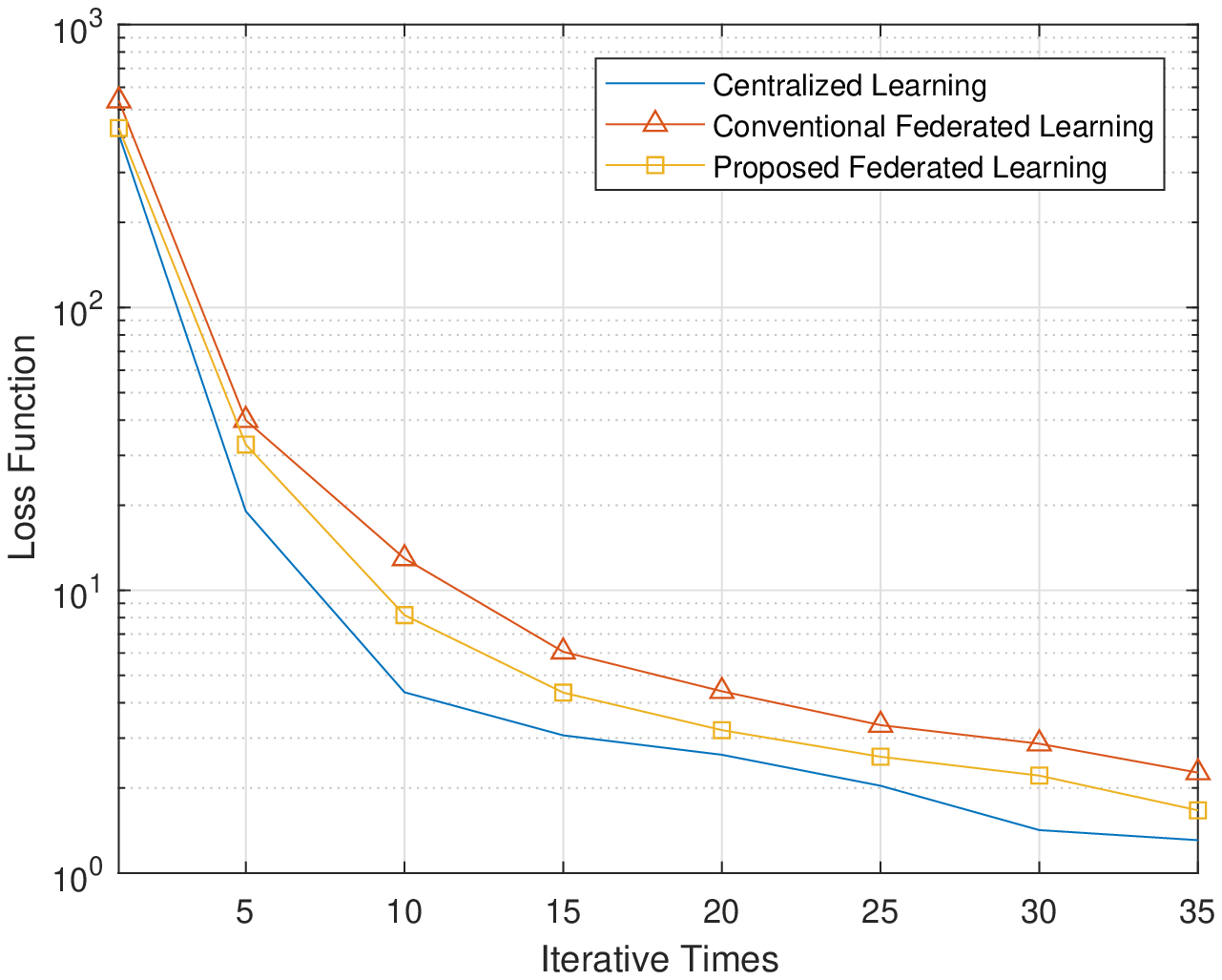}
\end{minipage}%
}%
\centering
\caption{The corresponding performance versus iterative times under expectation-based model.}
\label{exp_time}
\end{figure}

\begin{figure}[!t]
\centering
\subfigure[The accuracy performance versus number of nodes under expectation-based model.]{
\begin{minipage}[t]{0.45\textwidth}
\centering
\includegraphics[width=\textwidth]{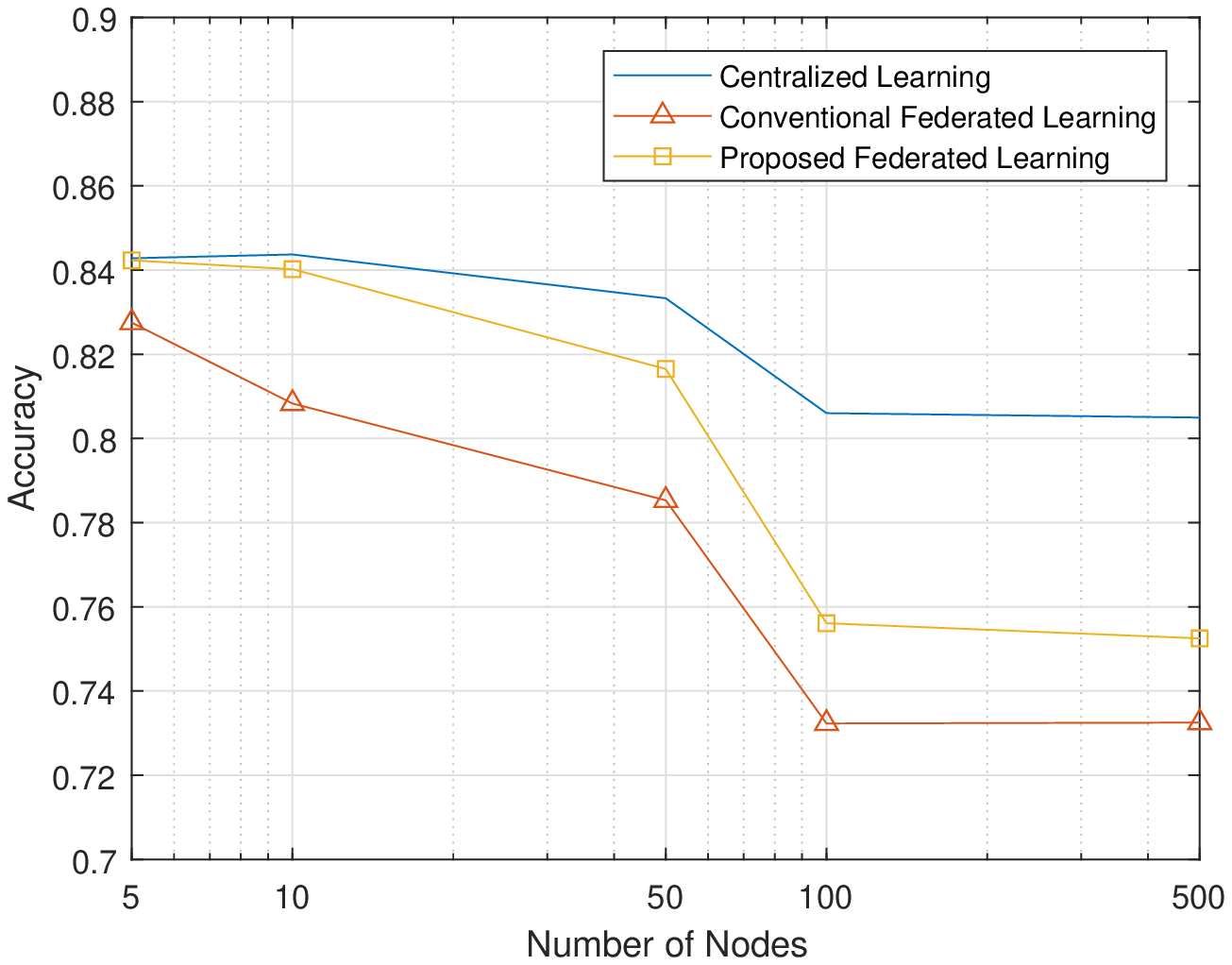}
\end{minipage}%
}%

\subfigure[The loss function performance versus number of nodes under expectation-based model]{
\begin{minipage}[t]{0.45\textwidth}
\centering
\includegraphics[width=\textwidth]{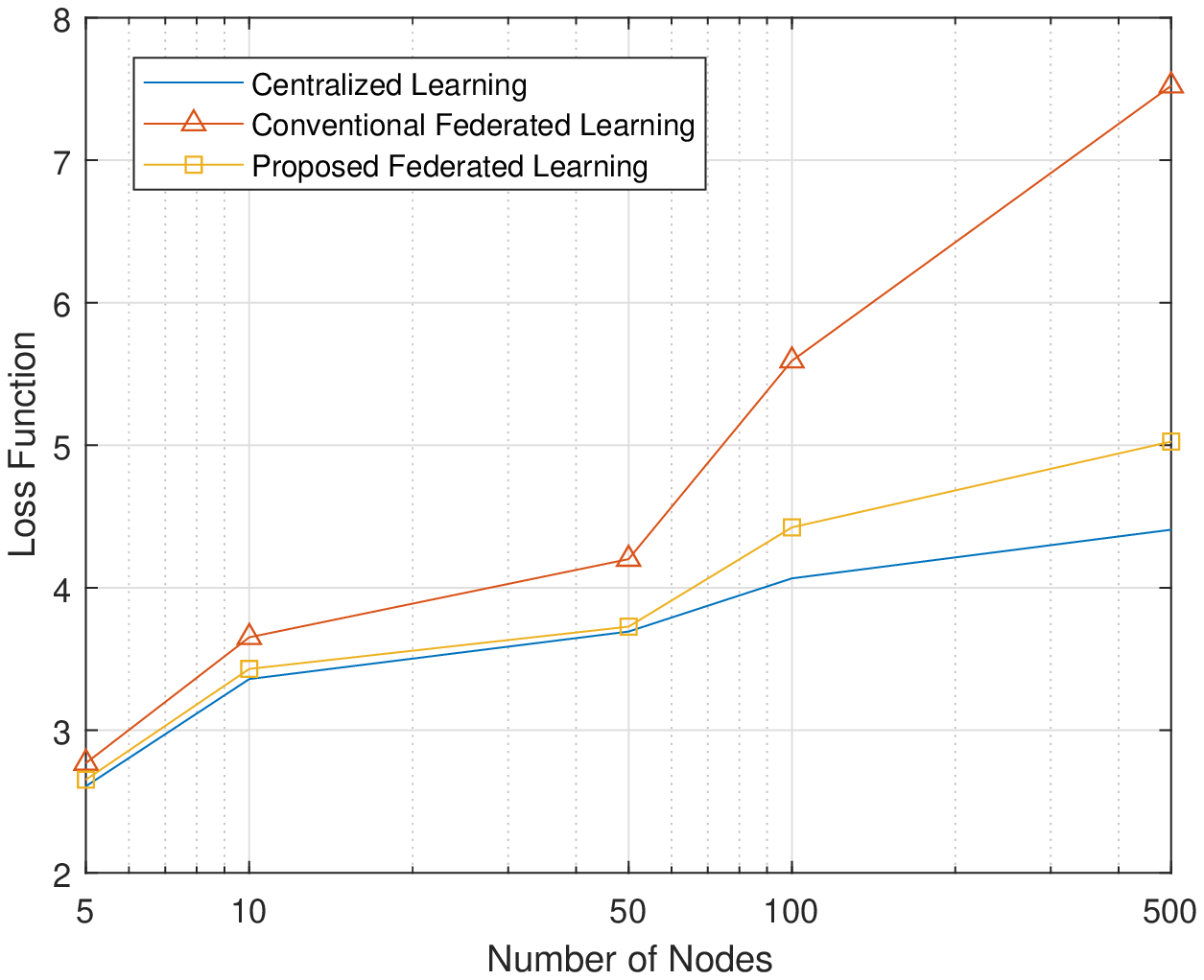}
\end{minipage}%
}%
\centering
\caption{The corresponding performance versus number of nodes under expectation-based model.}
\label{exp_nodes}
\end{figure}

In this section, we evaluate the performances of the proposed algorithms in image classification.
The simulation parameters are set as follows unless specified otherwise.
For illustration, we consider the learning task of image classification using the well-known MNIST dataset that consists of 10 categories ranging from digit ¡°0¡± to ¡°9¡± and a total of 70,000 data (60,000 for training and 10,000 for testing).
Besides, we exploit the SVM classification as our loss function for the training process, which outputs a binary label that corresponds to whether the digit is even or odd.
We consider the data partitions as i.i.d. in the distributed nodes, i.e., each data sample is randomly assigned to the nodes.
The performance is measured as the prediction accuracy and the values of loss function  with respect to the training dataset versus iteration count $t$.

For an intuitive comparison, we consider the following baseline approaches:
\begin{itemize}

\item[.]Centralized training, where the model is trained via a standard gradient descent procedure and the received value is estimated perfectly.

\item[.]Conventional federated training, which consists of the parameter noise and utilizes the imperfect estimated value to represent the real value for the training process. The model is trained via a standard gradient descent procedure and the loss function is the same as the centralized gradient descent.
\end{itemize}

\subsection{Simulations Under Expectation-based Model}

We set the noise variance for expectation-based model as $\sigma_e^2  = 1$.
We evaluate the prediction accuracy and the values of loss function as a function of iterations in Fig. \ref{exp_time}.
It agrees with our intuition that the noise has a serious impact on the learning model.
The prediction accuracy of the proposed algorithm is  higher than the conventional federated training.
The performance gap between two schemes increases with the iteration process as in Fig. \ref{exp_time}(a).
The result has a profound and refreshing implication that the added regularizer draws the model into a flat region so that the learning model has the ability to resist the noise.
In Fig. \ref{exp_time}(b), the values of loss function in the proposed scheme outperform the conventional method, which implies that our designed regularizer imposes appropriate punishment.

\begin{figure}[!t]
\centering
\subfigure[The accuracy performance versus iterative times under worst-case model.]{
\begin{minipage}[t]{0.45\textwidth}
\centering
\includegraphics[width=\textwidth]{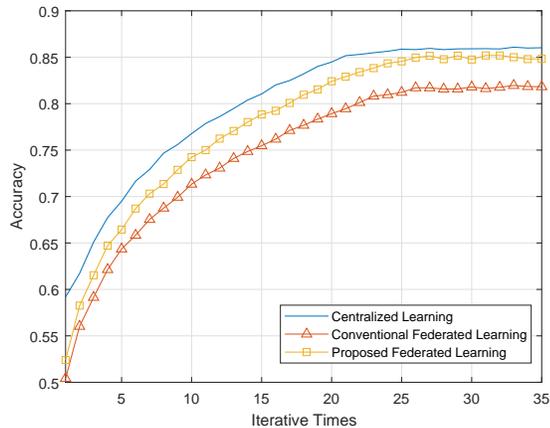}
\end{minipage}%
}%

\subfigure[The loss function performance versus iterative times under worst-case model.]{
\begin{minipage}[t]{0.45\textwidth}
\centering
\includegraphics[width=\textwidth]{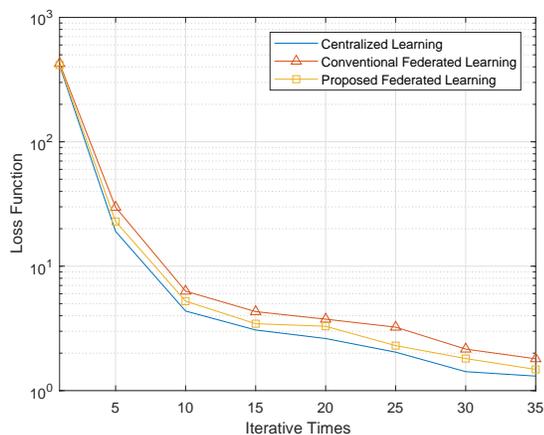}
\end{minipage}%
}%
\centering
\caption{The corresponding performance versus iterative times under worst-case model.}
\label{worst_time}
\end{figure}

\begin{figure}[t]
\centering
\subfigure[The accuracy performance versus number of nodes under worst-case model.]{
\begin{minipage}[t]{0.45\textwidth}
\centering
\includegraphics[width=\textwidth]{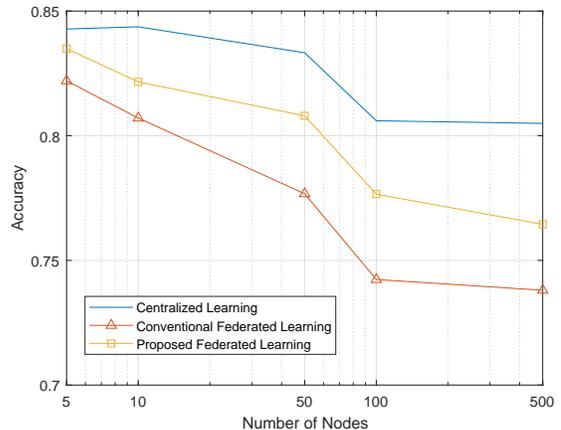}
\end{minipage}%
}%

\subfigure[The loss function performance versus number of nodes under worst-case model.]{
\begin{minipage}[t]{0.45\textwidth}
\centering
\includegraphics[width=\textwidth]{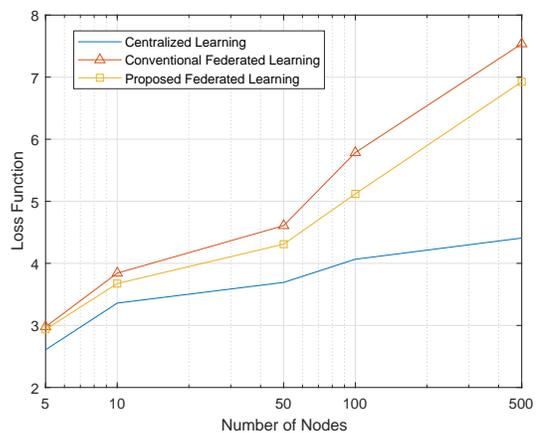}
\end{minipage}%
}%
\centering
\caption{The corresponding performance versus number of nodes under worst-case model.}
\label{worst_nodes}
\end{figure}

Fig. \ref{exp_nodes} shows the prediction accuracy and the values of loss function for the different numbers of nodes.
The observations show that the proposed algorithm has a better performance than the conventional scheme for both the prediction accuracy and the values of loss function.
Furthermore, we notice that the performance of the learning model decreases with the growth of the numbers of nodes, as we randomly divided the data samples to all nodes.
The division mode makes that each node has uniform information but not full information, and it will cause the hardship to find the optimal point during the training process.
With the growth of the numbers of nodes, each node can only obtain less and less samples and information, which leads to the decrease of the learning performance.
As shown in Fig. \ref{exp_nodes}(a), it is interesting to find that the accuracy of the proposed design is approaching to the centralized learning with few nodes, which proves the remarkable performance of the proposed design and verifies the direct effects of the proposed regularizer.
As illustrated in Fig. \ref{exp_nodes}(b), the values of loss function in the proposed design outperform than the conventional scheme, with the growth of the number of nodes especially.

\subsection{Simulations Under Worst-case Model}

We set the spherical region size of the noise for worst-case model as $\sigma_w^2 = 1$, and choose the sample noise sequence as $\|{\Delta {\bf{w}}} _j^t\|^2 = \sigma_w^2$, $j = 1, 2, ..., N$, $t = 0,1,2,...$.
Fig. \ref{worst_time} illustrates the prediction accuracy and the values of loss function for the different iterative times.
Without consideration of robust design, we notice that the noise reduces the accuracy performance of the training processes as shown in Fig. \ref{worst_time}(a).
The accuracy performance of the proposed scheme is significantly improved, which verifies that the added punishment of the loss function positively affects  the noise.
With the development of the iteration process, we can obtain that the performance of the proposed design approaches to the centralized training method.
The observations align with our discussions in Remark 3.
As shown at Fig. \ref{worst_time}(b), the values of loss function in all three schemes decrease with the iteration process.
It is interesting to see that the proposed scheme shows better performance, which proves the effectiveness of the approximation method.

We show the prediction accuracy and the values of loss function with different numbers of nodes in Fig. \ref{worst_nodes}.
With the increase of the number of nodes, the values of loss function and the prediction accuracy  of all designs are decreased.
However, the robust design of the proposed algorithm performs a remarkable gap in accuracy performance than the conventional design.
It is observed that the gap between the conventional training and the proposed design increases with the number of nodes as illustrated in Fig. \ref{worst_nodes}(a).
Such accurate learning of the proposed design is due to the positive punishment of the loss function and the proper approximation method which makes the global model robust against noise.
In Fig. \ref{worst_nodes}(b), the values of loss function in the proposed design outperform the conventional method,  and the gap is still increasing with the growth of the number of nodes.
This phenomenon verifies that the added punishment and the approximation method behave good effects on the training process.

\section{Conclusions}
In this paper, we have proposed the robust federated learning to resist the noise from  wireless communications.
Considering the noise in both aggregation and broadcast process, we have formulated the problem with effective noise as a parallel optimization problem under the expectation-based model and the worst-case model.
The corresponding optimization problem under the expectation-based model has been handled via the SLA algorithm, which can transform the effects of noise as  a designed regularizer in the loss function during the training process.
We have proposed the sampling-based SCA algorithm to solve the optimization problem under the worst-case model. Moreover, the convergent properties of both proposed designs have been derived that proposed designs have acceptable convergence rates.
Simulation results have illustrated that both proposed training processes under the mentioned models have improved the prediction accuracy and the values of loss function due to the proper punishment in the training.

\ifCLASSOPTIONcaptionsoff
  \newpage
\fi



%
%
%

\bibliography{robust_fed}
\bibliographystyle{IEEEtran}

\end{document}